\documentclass[sigconf, nonacm]{acmart}
\acmConference[GRAIL @ CIKM '26]{GRAIL 2026: Workshop on Generative, Retrieval-augmented, and Agentic Intelligence for Personalization}{November 8, 2026}{Rome, Italy}

\usepackage{booktabs}
\usepackage{microtype}
\usepackage{comment}
\usepackage{placeins} %
\usepackage{wrapfig}  %
\usepackage{balance} 

\usepackage{tikz}
\usepackage{graphicx}
\usepackage{stfloats}
\usetikzlibrary{shapes.geometric}
\usetikzlibrary{positioning,arrows.meta,decorations.pathreplacing}
\begin{document}
\settopmatter{printacmref=true}
\acmDOI{}
\acmISBN{}

\title{SIFT: Search Intent-to-Filter Transformer for Multi-Task Personalized Filter Ranking at Airbnb}

\author{Shashank Dabriwal}
\affiliation{\institution{Airbnb}\city{San Francisco}\country{USA}}
\email{shashank.dabriwal@airbnb.com}

\author{Tanya Piplani}
\affiliation{\institution{Airbnb}\city{San Francisco}\country{USA}}
\email{tanya.piplani@airbnb.com}

\author{Hao Li}
\affiliation{\institution{Airbnb}\city{San Francisco}\country{USA}}
\email{hao.li@airbnb.com}

\author{Yiwei Wang}
\affiliation{\institution{Airbnb}\city{San Francisco}\country{USA}}
\email{yiwei.wang@airbnb.com}

\author{Ashish Jain}
\affiliation{\institution{Airbnb}\city{San Francisco}\country{USA}}
\email{ashish.jain@airbnb.com}

\author{Kedar Bellare}
\affiliation{\institution{Airbnb}\city{San Francisco}\country{USA}}
\email{kedar.bellare@airbnb.com}

\author{Stephanie Moyerman}
\affiliation{\institution{Airbnb}\city{San Francisco}\country{USA}}
\email{stephanie.moyerman@airbnb.com}

\begin{abstract}

Search filters help guests navigate vast catalogs in two-sided marketplaces like Airbnb, and recommending the right filters can meaningfully lift booking conversion. Many such production filter-ranking systems, however, represent the guest through hand-engineered, pre-aggregated features generated by ETL pipelines. This makes it expensive to maintain and difficult to extend for new filter types or contextual dimensions (trip length, group size).

 We present \textbf{SIFT} (Search Intent-to-Filter Transformer), a ranking model built on transformers that learns guest preferences directly from raw behavioral sequences. SIFT replaces manual feature engineering with a unified guest representation that feeds multiple prediction tasks, including booking likelihood, filter engagement, and ordinal capacity thresholds (e.g., 2+ bedrooms) — a general framework for filter ranking in two-sided marketplaces that accommodates both boolean and numeric-range filter types. Extending SIFT to new filters requires only adding a new head, not a new feature pipeline. To keep serving fast, this guest representation is computed offline on a daily cadence rather than at request time.

Offline, SIFT improves booking and amenity-engagement PR-AUC by $+51.9\%$ and $+62.8\%$ respectively over the production baseline. In online A/B testing, SIFT increased engagement with recommended filters by $+20.0\%$, overall filter usage among searchers by $+0.72\%$, and usage of the newly-supported bedroom, bathroom, and bed filters by $+3.9\%$, $+10.7\%$, and $+0.52\%$ respectively. Demonstrating the system's extensibility, we rapidly integrated a novel hotel-intent filter using the same shared representation, driving a $+3.8\%$ lift in uncancelled hotel bookings and a $+0.76\%$ lift in overall marketplace bookings. SIFT is now fully deployed in production, serving scalable personalization to millions of guests.
\end{abstract}



\renewcommand{\shortauthors}{Dabriwal et al.}
\maketitle
\makeatletter
\fancypagestyle{standardpagestyle}{%
  \fancyhf{}%
  \renewcommand{\headrulewidth}{\z@}%
  \renewcommand{\footrulewidth}{\z@}%
  \fancyhead[LE,RO]{\@headfootfont\acmConference@shortname,
    \acmConference@date, \acmConference@venue}%
  \fancyhead[LO]{\@headfootfont\shorttitle}%
  \fancyhead[RE]{\@headfootfont\@shortauthors}%
}
\pagestyle{standardpagestyle}
\makeatother

\section{Introduction}
\label{sec:intro}

The Airbnb accommodation search exposes guests to millions of listings that
span various types of property, amenities, and price ranges. A guest's initial query
carries only coarse intent -- a destination, a date range, a party size---while
the preferences that actually determine which listings are acceptable often go
unstated. Search filters are the mechanism through which these latent
preferences become explicit, and previous work established that \emph{proactively}
recommending filters, rather than leaving their discovery to the guest, drives
incremental booking conversion of up to $+2.4\%$~\cite{li2026filters}. Filter
recommendation is therefore not a peripheral interface concern, but a
lower-funnel ranking problem on its own.

\begin{figure}[t]
  \centering
  \includegraphics[width=0.95\columnwidth]{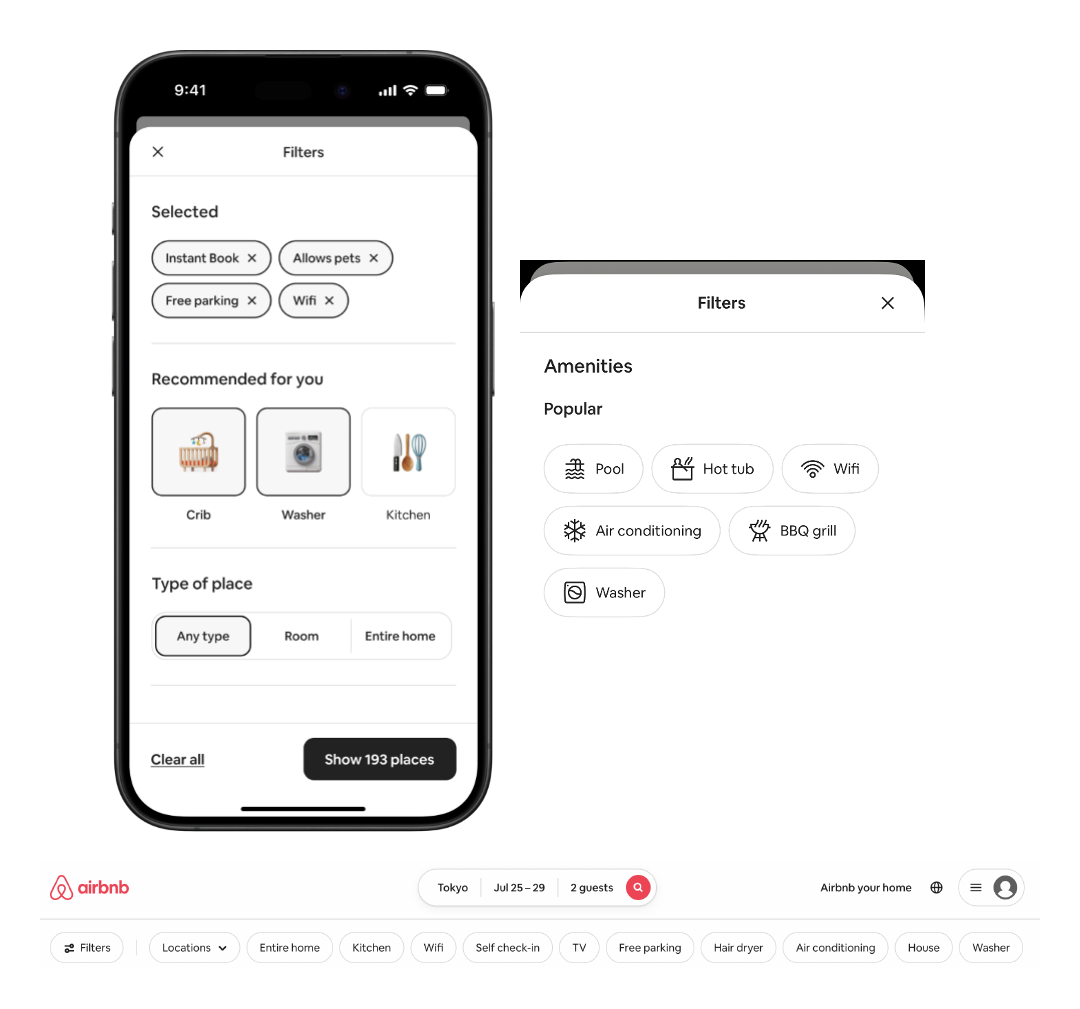}
  \caption{Recommending filters in ``Recommended for you'' and ``Amenities'' section in filter panel and beneath the search bar.}
  \Description{Screenshots of the Airbnb search interface showing recommended filter chips beneath the search bar and in the Recommended for you and Amenities sections of the filter panel.}
  \label{fig:ui-filter-common}
\end{figure}

The production system that established this result scores filters with a
two-layer MLP over hand-engineered, pre-aggregated user features -- counts such
as ``number of times a guest searched and then booked with a given filter over trailing 30 days'', computed by nightly batch
jobs. The design has served the surface well and remains inexpensive to
operate, but it has reached the limits of what an aggregate-feature
representation can express, and its two-head architecture covers only a subset
of the filters guests actually use. We pair each limitation below with the
mechanism SIFT introduces to address it.
 
\begin{itemize}
  \item \textbf{Feature brittleness and compute cost.} Every new filter type
  requires new features, and every new aggregate requires its own CPU-bound
  pipeline to author, backfill, and maintain. The same cost recurs even without a new filter: re-cutting an existing aggregate — such as the 30-day search-then-booked count above — along a new contextual dimension (e.g., trip length, group size) also requires a new pipeline. Expansion is gated by ETL
  throughput rather than by modeling capacity. SIFT removes this dependency: its sequence encoder consumes raw events and contextual attributes directly, so a new filter type or contextual dimension requires a new head or token feature, not a new pipeline.
 
  \item \textbf{Temporal blindness.} An aggregate collapses a behavioral
  sequence into a scalar. It cannot distinguish a guest's recent drift in
  preference from their multi-year average, and because it discards event order
  it cannot represent that a particular sequence of filtered searches
  culminated in a booking. SIFT's \emph{journey encoder}
  preserves order and recency and attends selectively over past events.
 
  \item \textbf{Ordinal collapse.} 
  The incumbent heads model booking likelihood and binary amenity engagement only; the numeric capacity filters (beds, bedrooms, bathrooms) are unmodeled entirely. Folding them into a multi-label engagement head would treat "2 bedrooms" and "5 bedrooms" as unrelated classes, discarding the ordering that makes them predictable and forcing each value to be learned independently. SIFT instead uses an ordinal decomposition that predicts a threshold ("at least two bedrooms") rather than an unordered class, sharing statistical strength across values via monotonically-constructed training labels.
 
  \item \textbf{Credit assignment.} A booking that follows a filtered search is
  not necessarily \emph{caused} by the filter. Treating all post-filter
  bookings as equally positive teaches the model to predict that a guest books,
  not that a filter helped them book, biasing the objective toward globally
  popular filters that merely co-occur with high-intent sessions. SIFT's
  \emph{attribution-discounted booking objective} reweights each booking by the
  search-result rank of the booked listing, upweighting conversions plausibly
  attributable to the filter's effect on the result set.
\end{itemize}
 
Sequence modeling over raw event histories is the established remedy for the
first two limitations, and at Airbnb the JourneyFormer
framework~\cite{journeyformer} already produces such journey-level guest
representations for accommodation search listing ranking. SIFT is, to our knowledge, the first
system to bring them to filter ranking; \S\ref{sec:related} situates this
against prior work on filter ranking and sequential user
modeling.
 
Concretely, SIFT trains a journey encoder end-to-end over the guest's
full event stream jointly with three types of discriminative task
heads that share the resulting representation. The filter-ranking
setting makes this factorization unusually favorable: the candidate
space is a small, fixed, fully enumerable filter vocabulary rather
than a catalog of millions, and---critically---only one of the three
task types depends on the candidate filter at all, engagement and
capacity being properties of $(\text{guest},\text{query})$ alone. We
develop how SIFT exploits this in \S\ref{sec:production}.
 
\medskip
\noindent
In summary, this work makes the following contributions.
 
\begin{itemize}
  \item \textbf{Framework.} We propose a personalized filter-ranking framework
  built on a single shared journey representation learned from raw events
  rather than maintained aggregates. It jointly optimizes binary booking
  prediction, multi-label amenity engagement, and ordinal prediction for
  numeric capacity filters, and ranks candidates by the product of booking and
  engagement probability, so that a recommended filter is judged by its
  contribution to conversion rather than by engagement alone. Because
  personalization is decoupled from query and filter signals, the framework
  extends to new filter types without new feature engineering---a property we
  validate in \S\ref{sec:hotel-intent} by extending it to a previously unmodeled
  hotel-intent class.
 
  \item \textbf{System.} We detail how SIFT is served within a search request's
  latency budget. \emph{Decoupled embedding materialization} keeps encoder
  inference off the request path entirely: a daily incremental batch job
  refreshes embeddings only for guests with new eligible events and writes them
  to a key-value store. \emph{Asymmetric head factorization} then scores the
  two query-conditioned head types once per request while scoring only the
  filter-conditioned booking head per candidate. We also address cold start,
  where guests with no history resolve to a \emph{learned cold-state embedding}
  rather than a hardcoded default one.
  
\item \textbf{Empirical validation.} SIFT improves booking PR-AUC by
$+51.9\%$ and amenity-engagement macro PR-AUC by $+62.8\%$ over the
production MLP baseline, and models three capacity filters the baseline
cannot express. Online, filter engagement rises substantially ($+20.0\%$
Recommended Filter Clickers, $+0.72\%$ Searchers With Filters overall,
and $+10.7\%$, $+3.9\%$, $+0.52\%$ for Bathroom, Bedroom, and Beds filter
usage respectively) and the low-inventory guardrail improves. Extended to
a new hotel-intent filter signal that targets \emph{Hotel Megaphone} --- a
promotional banner on the search page offering hotel credits to guests
with inferred hotel intent --- the same framework lifts uncancelled hotel
bookings by $+3.8\%$ online and overall marketplace bookings by $+0.76\%$.
\end{itemize}
 
While this study is grounded in Airbnb, the underlying problem---matching
diverse inventory to needs inferred from a user's history, in service of a
lower-funnel objective such as booking---is common to marketplaces that log
user behavior and expose many filters spanning heterogeneous intents. The
mechanisms we describe are correspondingly general.
 
The rest of the paper is organized as follows. Section~\ref{sec:related}
reviews related work. Section~\ref{sec:motivation} motivates the shift from
aggregated features to sequence modeling. Section~\ref{sec:formulation} formalizes the filter-ranking problem and
its three sub-tasks. Section~\ref{sec:model} presents SIFT's architecture.
Section~\ref{sec:training} describes training data construction, label
attribution, and cold-start handling. Section~\ref{sec:production} covers the
serving system. Section~\ref{sec:experiments} reports offline results,
online A/B results, an ablation study, and the hotel-intent generalization study.
Section~\ref{sec:discussion} discusses limitations, and
Section~\ref{sec:conclusion} concludes.

\section{Related Work}
\label{sec:related}

We discuss related work from two key areas of focus: 1) faceted and filter-based search, and 2) sequence-based recommendation models.

\paragraph{\textbf{Faceted and filter-based search}}
Facets have been studied for two decades as an interface for exploratory
information seeking, where the central question is which facets to expose, and
in what order, so that users can navigate a large collection without
formulating a precise query~\cite{hearst2006clustering, tunkelang2009faceted}. A
substantial line of work formalizes facet selection as an optimization over the
current result set: dynamic faceted retrieval selects facets that partition the
candidate set informatively at
scale~\cite{benyitzhak2008beyond}, while effort- and cost-driven formulations
select the facet sequence that minimizes a user's expected navigational cost to
reach a target item~\cite{roy2008minimum, kashyap2010facetor}. What unites these
approaches is that the objective is a property of the result set and of the cost
of traversing it; the identity of the searcher does not enter the formulation,
and two users issuing the same query receive the same facets. Recent work recast
filter selection at Airbnb as a supervised learning problem optimized against
downstream conversion rather than navigational
efficiency~\cite{li2026filters}---the direct baseline, whose representational and
coverage limits \S\ref{sec:intro} details. SIFT departs from both traditions by
conditioning the scoring function on the guest's accumulated behavioral history:
two guests issuing an identical query over an identical result set receive
different filter slates. Filter selection becomes a personalization problem
rather than a result-set summarization problem, and the relevant question shifts
from which facets best partition the candidates to which filters this guest is
likely to apply and to convert under.
 
\paragraph{\textbf{Sequence-based recommendation models}}
SASRec~\cite{kang2018self} showed that causal self-attention over a user's
interaction history outperforms Markov-chain and convolutional sequence models,
and BERT4Rec~\cite{sun2019bert4rec} obtained further gains from bidirectional
encoding under a masked-item objective. The consistent finding is that learned
sequence representations recover order and recency structure that pre-aggregated
features discard by construction -- the property that motivates replacing the
baseline's aggregate features (\S\ref{sec:intro}). Industrial ranking systems
have since adopted sequence encoders as components of larger
architectures~\cite{haldar2019applying, haldar2020improving}, and Airbnb's JourneyFormer
framework~\cite{journeyformer} scales the idea to journey-level modeling,
encoding a long-horizon sequence of infrequent high-intent events alongside a
short-horizon sequence of recent page views and compressing both into a single
guest embedding.

\section{Motivating Analysis}
\label {sec:motivation}
\paragraph{\textbf{Temporal Drift in Filter Preferences}}

We first measure how much an individual
guest's filter preferences actually change over time, using
production search logs spanning a range of trailing aggregation windows of
filtered search and booking activity. Naive comparisons across
different-length windows are confounded by window overlap and
seasonality, so we isolate two fully disjoint, season-matched
twelve-month periods per guest and compare filter usage between them.
All results below are robust to excluding automated/bot traffic and to
restricting to filters with stable population-level volume across
both periods.

A majority of guests' single most-used filter changes between the two
disjoint years (Table~\ref{tab:drift}), and this holds even when
looking at a guest's entire top-5 filter set rather than just their
single top choice.

\begin{table}[t]
\centering
\caption{Guest-level filter preference drift between two disjoint,
season-matched twelve-month periods.}
\label{tab:drift}
\begin{tabular}{@{}lr@{}}
\toprule
Metric & Value \\
\midrule
Top-1 filter disagreement rate & 60--65\% \\
Avg.\ top-5 filter set overlap & 53\% (2.65 / 5) \\
Guests sharing $\leq$2 of top-5 filters & 44\% \\
\bottomrule
\end{tabular}
\end{table}
\FloatBarrier

This instability is specific to \emph{which} filter a guest prefers,
not to how well that filter converts once applied. Population-level
Filter Conversion Rate (FCR) --- the share of searches with a given
filter that end in a booking --- moves far less than filter preference
does across the same two years (Table~\ref{tab:fcr}). Together, these
findings show that engagement preference changes far more rapidly than
conversion value does, and even conversion value's residual
instability is concentrated in a subset of low-volume filters rather
than spread evenly (the unweighted average change, 35.5\%, exceeds the
volume-weighted average, 24.2\%, and the largest single swing observed
is 88.3\%). A static, fixed-window aggregate has no mechanism to track
either kind of drift at the granularity an individual guest requires,
which motivates a sequence encoder that can adapt to both.
\begin{table}[b]
\centering
\caption{Population-level FCR drift over the same two disjoint years,
measured as $|$relative FCR change$|$ per filter.}
\label{tab:fcr}
\begin{tabular}{@{}lr@{}}
\toprule
Metric & Value \\
\midrule
Volume-weighted average & 24.2\% \\
Unweighted average       & 35.5\% \\
Maximum observed          & 88.3\% \\
\bottomrule
\end{tabular}
\end{table}

Individual cases illustrate the mechanism: several guests' top
ordinal capacity filter persists in type but shifts to a larger
threshold across years (e.g.\ a minimum-bedroom or minimum-bed filter
increasing) --- consistent with a growing travel party --- while
others show a filter newly entering their top-used set with no prior
usage at all, such as the \texttt{crib} amenity filter becoming a
guest's most-used filter after a year of zero use. Fixed-window
aggregates can only detect such a shift by manually differencing two
separately-maintained windows, and even then only learn that usage
changed \emph{sometime} within the year, not how recently. A sequence
model conditioned on ordered event history recovers that recency
directly, distinguishing a guest whose crib searches started three
days ago from one whose searches have been steady for ten months.

\section{Problem Formulation}
\label {sec:formulation}
Let $u$ denote a guest, $q$ a search query (destination, dates, guest count,
etc.), and let $f$ denote a candidate \emph{filter state}: the multi-hot vector of
31 boolean amenity flags (e.g., \textsc{Pool}, \textsc{Wifi}), and
numeric range filters $f_k \in \{$\textsc{Beds}, \textsc{Bedrooms},
\textsc{Bathrooms}$\}$ that take integer values from 1 to $K_k$.

We wish to learn a scoring function $s(u, q, f)$ that ranks each candidate
filter $f \in \mathcal{F}$ by its expected utility—i.e., its contribution to
booking conversion if surfaced.  We decompose this into three sub-tasks:

\begin{itemize}
  \item \textbf{T1 — Booking}: $P(B{=}1 \mid u, q, f)$, the probability that
        the session results in a booking when filter $f$ is applied.
  \item \textbf{T2 — Engagement}: $P(E_f{=}1 \mid u, q)$ for each amenity
        filter $f$, whether the user applies filter $f$ at all.
  \item \textbf{T3 — Ordinal value}: $P(V_k \geq v \mid u, q)$ for each
        threshold $v \in \{1, \ldots, K_k\!-\!1\}$ of numeric filter $k$,
        capturing the minimum acceptable value (e.g., at least 2 bedrooms).
\end{itemize}

The final ranking score multiplies the booking head's output with the
corresponding engagement probability: for a boolean amenity filter,
$P(\text{engage with } f \mid u, q) = P(E_f{=}1 \mid u, q)$ directly
from T2; for a threshold $v$ of numeric filter $k$ (e.g.,
\textsc{beds} $\geq 2$), the exact-value probability is recovered from
T3's adjacent cumulative outputs by subtraction: 
\[
P(\text{engage with
} f \mid u, q) = P(V_k \geq v \mid u, q) - P(V_k \geq v{+}1 \mid u,
q).
\]
The final ranking score is then
\[
  s(u, q, f) = P(B{=}1 \mid u, q, f) \cdot P(\text{engage with } f \mid u, q)
\]
so that a filter is ranked highly only if the guest is both likely to
apply it and likely to book as a result.

\section{Model Architecture}
\label{sec:model}

\begin{figure*}[t]
\centering
\includegraphics[width=0.72\textwidth]{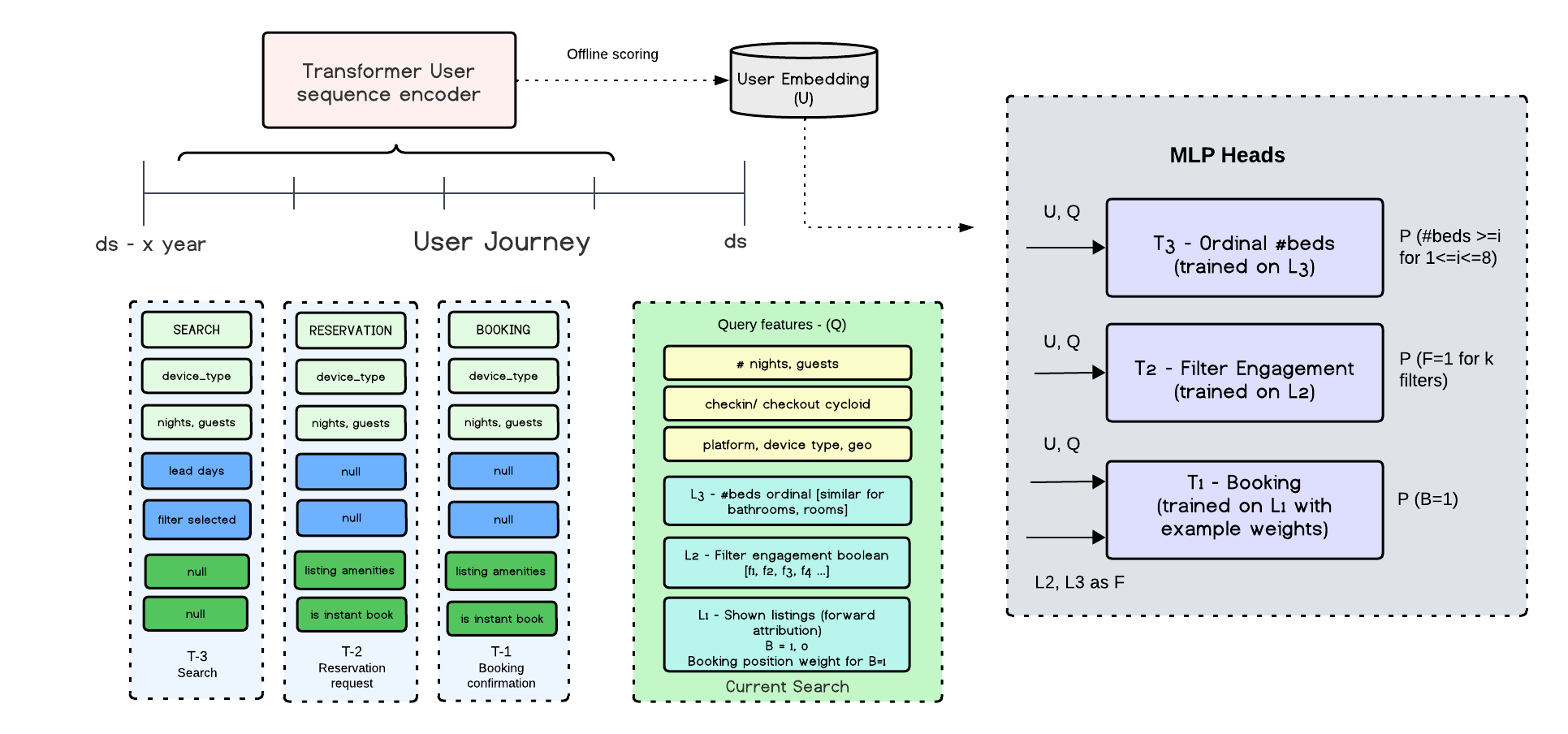}
\caption{SIFT training data construction and multi-task architecture. Each
guest's historical actions --- search, reservation-request, and
booking-confirmation events, each tokenized from discrete, null-padded features --- form a time-ordered sequence spanning the guest's
journey up to the current date, consumed by a transformer encoder
that produces a user embedding $U$ (i.e.\ $\mathbf{e}_u$), computed offline
in a daily scoring job and cached in an embedding store. At inference and
training time, $U$ is combined with the current search session's query
features $Q$, along with the candidate filter's label context $L_2$ (multi-hot engagement) and $L_3$ (ordinal capacity) --- to feed three types of task-specific MLP heads: $T_1$
(booking probability $P(B{=}1)$, trained on forward-attributed booking
labels $L_1$ with position-discount example weights), $T_2$ (per-filter
engagement probability $P(F_k{=}1)$, trained on $L_2$), and $T_3$ (one
ordinal head per numeric capacity filter, each predicting
$P(\text{\#beds} \geq i)$-style thresholds, trained on $L_3$).}
\Description{Diagram of the SIFT pipeline: a time-ordered sequence of guest search, reservation-request and booking events feeds a transformer encoder that outputs a user embedding, which is combined with query features and candidate filter context and passed to booking, engagement and ordinal capacity MLP heads.}
\label{fig:sift-pipeline}
\end{figure*}

Figure~\ref{fig:sift-pipeline} illustrates the overall pipeline: a transformer
\emph{SIFT Encoder} consumes a guest's raw event sequence to produce a user
embedding, materialized daily; a \emph{Feature Assembly} layer combines that
embedding with the current query and candidate filter features; and three
types of \emph{Task-Specific MLP Heads} ($T_1$ booking, $T_2$ engagement, $T_3$
ordinal capacity) score the result.

\subsection{Architecture}\label{sec:architecture}
A transformer encoder, trained jointly end-to-end with
  all three head types, consumes a single chronological sequence of a guest's past
  searches (with filter applied and other context) and reservation related events. Each token's raw
features are first summed with a standard sinusoidal positional
encoding~\cite{vaswani}, then projected into the encoder's model
dimension via a compression MLP
(1024$\rightarrow$\allowbreak512$\rightarrow$256$\rightarrow$\allowbreak128, swish
activations) before self-attention. It uses a
  single self-attention layer (4 heads, head dimension 32, model dimension $d{=}128$). The encoder output is compressed by
a further MLP projection (512{\textrightarrow}128{\textrightarrow}32) into a
contextual user embedding $\mathbf{e}_u\in\mathbb{R}^{32}$ taken from the
final sequence position. This embedding feeds three
types of task-specific two-layer MLP heads with swish activations: a
booking head, an amenity-engagement head, and an ordinal head for each
numeric capacity filter (beds, bedrooms, bathrooms). The ordinal heads
use the Frank \& Hall ordinal decomposition~\cite{frankball} ($K{-}1$
binary sub-classifiers per filter, illustrated in Figure~\ref{fig:ordinal-labels}). The model is trained with a joint
multi-task loss (\S\ref{subsec:heads}) using
position-discount sample weights $w_i=1+1/\ln(\mathrm{pos}_i+1)$ applied
  to the booking head's loss, where $\mathrm{pos}_i$ is the rank of the booked
  listing in the search results returned after the candidate filter was
  applied---upweighting bookings more plausibly attributable to the filter
  itself rather than chance (Figure~\ref{fig:position-discount}). The full model
  with the encoder and all three head types is trained jointly end-to-end with the Adam
  optimizer ($\beta_1{=}0.9$, $\beta_2{=}0.98$, $\epsilon{=}10^{-9}$) under a Transformer-style warmup + inverse-square-root decay learning-rate
  schedule, batch size 512, using
  Horovod-distributed data-parallel training on 20 GPUs in
  ${\approx}12$~hours. Figure~\ref{fig:sift-pipeline} shows SIFT's complete training data construction and multi-task architecture.

\subsection{SIFT Encoder (User Sequence Modeling)}
\label{subsec:gr}

SIFT's encoder consumes a guest's raw event history --- searches with
filters applied, reservation requests, booking confirmations, and cancellations --- as a
single chronological sequence. These event types carry different,
only partially-overlapping sets of attributes, so a single encoder
cannot consume them directly without first putting them into a common
representation. We adopt a PinnerFormer-style~\cite{pinnerformer} token
schema: every event, regardless of type, is encoded into the same
fixed-width dense feature vector, composed of shared segments (an
action-type indicator; platform and device type; and trip context such
as check-in/check-out dates --- cyclically encoded via sine/cosine
transforms of month, day-of-month, and day-of-week to preserve
periodicity across year/month/week boundaries -- and guest count) followed by event-type-specific
segments (search-specific: destination, lead time, filter values applied; reservation-specific: instant-book flag, and listing attributes such as capacity, amenities). Segments that do not apply to a given
event's type are null-padded rather than omitted, so every token
occupies an identical schema and can be consumed by a single shared
encoder without a per-action-type input branch. Recency is captured
through positional encoding and event order. The transformer self-attention mechanism
learns to weight historically relevant events when producing the
fixed-size user embedding $\mathbf{e}_u \in \mathbb{R}^d$.

\subsection{Feature Assembly}
\label{subsec:features}

The user embedding $\mathbf{e}_u$ and query features $\mathbf{q}$ are
common inputs shared by all three head types. The booking head
additionally receives candidate filter features $\mathbf{f}_c$, since
$P(B{=}1 \mid u, q, f)$ depends on which specific filter is being
scored, while the engagement and ordinal heads do not:
\begin{equation}
  \mathbf{x}_{T1} = [\mathbf{e}_u \;\|\; \mathbf{q} \;\|\; \mathbf{f}_c],
  \qquad
  \mathbf{x}_{T2} = \mathbf{x}_{T3} = [\mathbf{e}_u \;\|\; \mathbf{q}]
\end{equation}

Query features include destination geo embedding, number of nights,
number of guests, lead time, cycloid checkin/checkout times, and
device/platform indicators. Candidate filter features $\mathbf{f}_c$
are a multi-hot vector over the boolean amenity flags together with
the candidate ordinal threshold values for beds, bedrooms, and
bathrooms, identifying which specific filter is being evaluated as a
candidate.

This design \emph{decouples} personalization (captured entirely by
$\mathbf{e}_u$) from query and filter signals, and removes the need for dozens of
pre-aggregated user feature pipelines (e.g., 30/90/365-day filter-usage
histograms) that were used previously.

\subsection{Task-Specific MLP Heads}
\label{subsec:heads}

Each task head is a two-layer MLP with swish activations applied to its corresponding
representation $\mathbf{x}_{T1}$, $\mathbf{x}_{T2}$, or $\mathbf{x}_{T3}$
(\S\ref{subsec:features}).

\begin{figure}[t]
\centering
\includegraphics[width=\columnwidth]{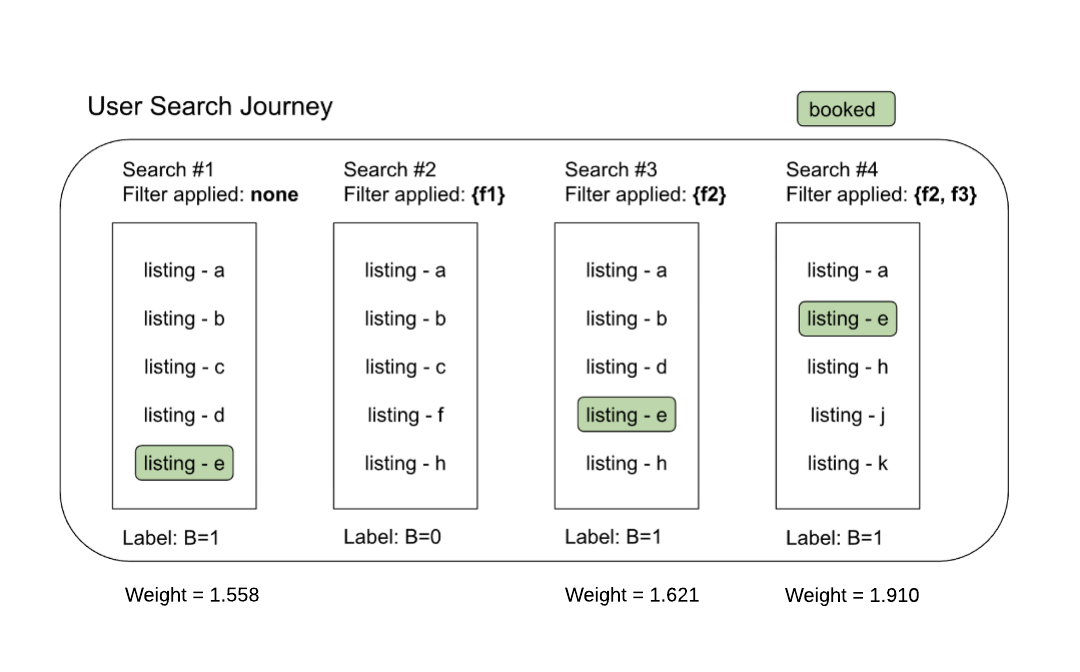}
\caption{Position-discount example weighting for the booking objective.
Across a guest's search journey, the same listing (\textsc{e}) is booked
after appearing at different ranks depending on which filters were
applied: 5th with no filter (Search~\#1, $w{=}1.558$), not shown at all
with filter \{f1\} (Search~\#2, $B{=}0$), 4th with filter \{f2\}
(Search~\#3, $w{=}1.621$), and 2nd with filters \{f2, f3\}
(Search~\#4, $w{=}1.910$), using $w_i = 1 + 1/\ln(\mathrm{pos}_i+1)$.
Bookings following a search where the listing ranked higher receive more
weight, since they are more plausibly attributable to the filter's effect
on the result set rather than the guest scrolling past many alternatives.}
\Description{Four example searches from one guest journey, each with a different filter set, showing the rank at which the booked listing appeared and the resulting position-discount weight.}
\label{fig:position-discount}
\end{figure}

\paragraph{\textbf{T1 — Booking Head}}
A sigmoid output predicts $\hat{p}_B = P(B{=}1 \mid u, q, f)$ using standard
binary cross-entropy (BCE):
\begin{equation}
  \mathcal{L}_B = -\sum_{i} w_i \bigl[ y_i \log \hat{p}_i + (1{-}y_i)
  \log(1{-}\hat{p}_i) \bigr]
\end{equation}
where the position-discount weight is
$w_i = 1 + 1/\ln(\text{pos}_i + 1)$,
giving more credit to bookings that occurred with the booked listing near the
top of search results (Figure~\ref{fig:position-discount}).

\begin{figure}[t]
\centering
\includegraphics[width=\columnwidth]{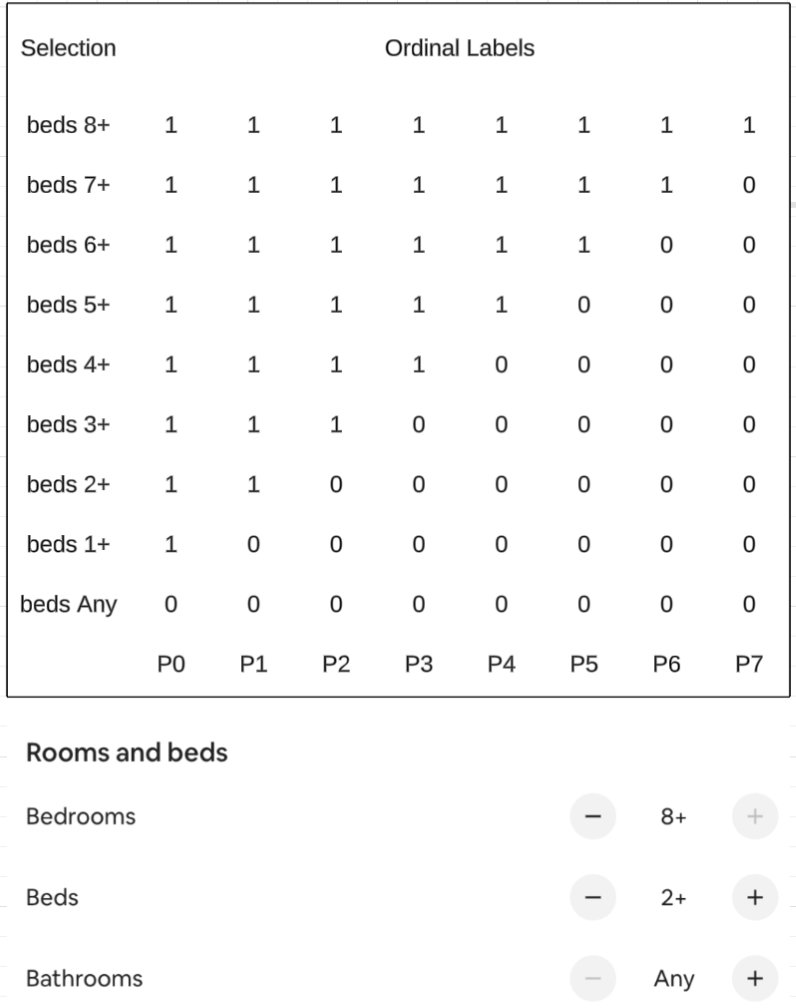}
\caption{Frank \& Hall label construction for the \textsc{Beds} filter
($K{=}9$ levels, $K{-}1{=}8$ sub-classifiers P0--P7). A guest's applied
selection (e.g.\ \textsc{beds 2+}) expands into a monotone binary vector
$\mathbf{1}[V_k^* \geq v]$ for each threshold $v$.}
\Description{Grid showing how a guest's Beds 2+ selection maps to a monotone binary label vector across eight threshold sub-classifiers P0 to P7.}
\label{fig:ordinal-labels}
\end{figure}

\paragraph{\textbf{T2 — Amenity Engagement Head}}
A multi-label sigmoid layer with one output per amenity filter predicts
$\hat{p}_{E_f}$ independently.  The loss is the sum of per-label BCE losses
over the amenity vocabulary.
\paragraph{\textbf{T3 — Ordinal Value Heads}}
For each numeric filter $k$ with $K_k$ discrete levels we follow the Frank \&
Hall~\cite{frankball} decomposition and train $K_k{-}1$ binary
sub-classifiers, each predicting $P(V_k \geq v)$ (Figure~\ref{fig:ordinal-labels}),
using the same independent multi-label mechanism as T2. Because the
training labels $\mathbf{1}[V_k^* \geq v]$ are monotonically
non-increasing in $v$ by construction, predicted probabilities tend to
be monotonic in practice as well. The loss for filter $k$ is:
\begin{equation}
  \mathcal{L}_{O_k} = \sum_{v=1}^{K_k-1}
    \text{BCE}\!\left(\hat{P}(V_k \geq v),\; \mathbf{1}[V_k^* \geq v]\right)
\end{equation}
where $V_k^*$ is the value the user applied in the session.

\paragraph{\textbf{Joint Loss}}
\begin{equation}
  \mathcal{L} = \lambda_B \mathcal{L}_B +
                \lambda_E \mathcal{L}_E +
                \sum_{k} \lambda_{O_k} \mathcal{L}_{O_k}
\end{equation}
Weights $\lambda$ are tuned as hyperparameters; in our experiments
$\lambda_B{=}1.0$, $\lambda_E{=}10.0$, $\lambda_{O_k}{=}1.0$ for all $k$.

\section{Training}
\label{sec:training}

\subsection{Training Data and Label Construction}\label{sec:training-data}

The training corpus is built from a rolling 365-day
  window of daily-generated user journeys, each ending in a 1-day snapshot of
  search events with 7-day forward-attributed booking labels. Each
  journey's historical sequence is drawn from booking events and searches with filters applied collected over multiple years, truncated to 384 tokens.  Each training instance
corresponds to a single search session and carries:
\begin{itemize}
  \item The guest's historical journey which will be used to generate SIFT user embedding $\mathbf{e}_u$ 
  \item Query features extracted from the search record.
  \item For T1: a binary booking label and the position-discount weight of the
        booked listing.
  \item For T2: multi-hot amenity labels indicating which filters the guest
        applied in this session.
  \item For T3: the monotonic binary vector $\mathbf{1}[V_k^* \geq v]$ as label
        derived from the numeric value (beds, bedrooms, bathrooms) the
        guest applied as a range filter, if any.
\end{itemize}

To avoid session-level label leakage, we first collapse each session to
a single representative search event, preferring one with filter
interactions when multiple exist. We then apply a forward-attribution
scheme, scanning forward from that retained event within a 7-day window:
a booking is attributed to the search if the booked listing ID appears
among the listings shown in that search's result set. This avoids
crediting every search event in a session with an eventual booking when
only one of them plausibly drove it.

\subsection{Cold-Start Optimization}\label{sec:cold-start}
Guests with no prior history are
  represented by a single \texttt{UNKNOWN\_\allowbreak ACTION} placeholder token, which
  falls outside the event-type vocabulary and resolves to a dedicated
  out-of-vocabulary embedding. Unlike a hardcoded zero vector, this embedding
  is learned jointly with the rest of the model, giving zero-history guests a
  data-driven cold-state representation rather than an arbitrary default.

\section{System Design and Productionization}
\label{sec:production}
  \subsection{Serving}\label{sec:serving}
  SIFT splits inference across an offline and an online stage. Offline, the
  exported encoder is run daily over each guest's latest event journey,
  producing a refreshed contextual user embedding $\mathbf{e}_u$ that is
  written to an online key-value store; only guests with a new eligible event
  since the last run require recomputation. Online, a guest search request
  triggers filter scoring online, before search results are returned: the
  guest's precomputed offline embedding is fetched from the store and combined with
  online query features, avoiding any encoder inference on the serving path.
  Since the booking head requires the candidate filter state $f$ as input,
  it is scored per candidate filter, returning a single probability per
  filter. The engagement and ordinal heads, by contrast, depend only on
  $(u,q)$ and are therefore scored once per query, returning a vector of
  probabilities spanning all candidate filters in a single forward
  pass---an efficiency benefit of the shared-encoder, multi-head design over a
  per-filter scoring architecture. 
  \subsection{Latency and Throughput}
SIFT targets the same tight latency budget as the predecessor system
on this surface~\cite{li2026filters}: sub-10ms inference latency at
over 1000 QPS. Model serving uses a sidecar architecture
pattern~\cite{burns2016sidecar}, co-locating a TensorFlow Serving
sidecar with the search service in the same pod and eliminating the
network hop a remote serving call would otherwise add. Feature
serving is shared with the listing search module rather than
duplicated per module: a single feature fetch reduces latency, shared
feature computation avoids redundant work across modules, and
centralized feature management keeps SIFT's inputs consistent with
listing search's. SIFT meets this budget despite its heavier transformer encoder because of the offline/online split described in \S\ref{sec:serving} — the same pattern the underlying journey-encoder framework~\cite{journeyformer} uses to keep its own real-time inference lightweight — leaving only the three MLP heads to run per request within the sidecar.
\FloatBarrier

\section{Experiments}
\label{sec:experiments}
\subsection{Offline Evaluation}
We evaluated SIFT using PR-AUC, comparing against the production MLP baseline trained on hand-engineered features. For the multi-label engagement head, we report both macro PR-AUC (the average of per-filter PR-AUC scores, weighting every filter equally) and micro PR-AUC (a single PR-AUC computed by pooling all guest-filter predictions together, so common filters dominate); the three ordinal capacity filters (bedrooms, beds, bathrooms) are reported as the analogous macro PR-AUC, averaged across each filter's threshold sub-classifiers.

Table~\ref{tab:offline} compares SIFT against the baseline on the same held-out window. SIFT improves booking PR-AUC by $+51.9\%$ and amenity-engagement PR-AUC by $+62.8\%$ (macro) and $+35.5\%$ (micro), and additionally reports PR-AUC for the three ordinal capacity filters (bedrooms, beds, bathrooms), which the baseline cannot express at all since it has no mechanism to model numeric filter values. \S\ref{sec:ablation} isolates how much of this gain is attributable to the sequence encoder itself.

\begin{table*}[t]
\small
\centering
\caption{Offline PR-AUC comparison (held-out 7-day window) and journey-encoder ablation. Baseline is production MLP with
hand-aggregated features. Bedrooms/Beds/Bathrooms report macro PR-AUC across threshold sub-classifiers.}
\label{tab:offline}
\begin{tabular}{lcccccc}
\toprule
\textbf{Model} & \textbf{Booking} & \textbf{Eng. (Macro)} & \textbf{Eng.(Micro)} & \textbf{Bedrooms} & \textbf{Beds} & \textbf{Bathrooms} \\
\midrule
MLP Baseline    & 0.1388 & 0.1186 & 0.2875 & — & — & — \\
SIFT (ours)    & \textbf{0.2108} & \textbf{0.1931} & \textbf{0.3896} & \textbf{0.2349} & \textbf{0.1667} & \textbf{0.1570} \\
\midrule
\textit{Rel.~gain} & $+51.9\%$ & $+62.8\%$ & $+35.5\%$ & \textit{new} & \textit{new} & \textit{new} \\
\midrule
\multicolumn{7}{l}{\textit{Ablation}} \\
SIFT w/o journey encoder & 0.1410 & 0.1168 & 0.2572 & 0.1529 & 0.0676 & 0.0504 \\
\bottomrule
\end{tabular}
\end{table*}
\FloatBarrier

Analysis confirms all three head types are well calibrated: the ratio of mean predicted probability to mean observed label rate stays close to 1.0 per class, with no systematic over- or under-prediction, making the raw head outputs directly usable as ranking scores at serving time without additional recalibration.

\subsection{Online A/B Tests}

We ran a series of online A/B experiments on Airbnb's search surface to measure the business impact of SIFT against the production MLP baseline, each using a large randomized traffic split, with randomization done by identity (visitor or user ID) to ensure a consistent experience for both logged-out and logged-in traffic. Primary metrics are filter engagement metrics — searches with any filter applied, unique guests who applied any filter, and unique guests who applied each of the newly-supported bedroom, bathroom, and bed filters specifically; bookings and low-inventory search rate — the share of guests whose search returns too few results, which could be impacted by recommending filters that narrow results too aggressively — are tracked as guardrails. Table~\ref{tab:ab} presents the results.

\begin{table}[b]
\small
\caption{Online A/B results vs.\ MLP baseline. Statistically significant unless noted.}
\label{tab:ab}
\begin{tabular}{ll}
\toprule
\textbf{Metric} & \textbf{Relative lift} \\
\midrule
\textbf{Recommended Filter Clickers}      & $\mathbf{+20.0\%}$ \\
Searchers With Filter — Bathrooms  & $+10.7\%$ \\
Searchers With Filter — Bedrooms   & $+3.9\%$ \\
Searchers With Filter — Beds       & $+0.52\%$ \\
\textbf{Searches With Filters}    & $\mathbf{+1.1\%}$ \\
\textbf{Searchers With Filters}   & $\mathbf{+0.72\%}$ \\
Searchers Low Inventory (guard.)    & $-0.27\%$ (improved) \\
Bookings (guard.)                  & $+0.03\%$ (not stat-sig) \\
\bottomrule
\end{tabular}
\end{table}

\subsection{Ablation Study}
\label{sec:ablation}

We isolate the contribution of the design choices by retraining SIFT
with one component removed or substituted at a time, holding the
training window (the same period as Table~\ref{tab:offline}), optimizer,
and all other hyperparameters fixed.

Removing the journey encoder -- scoring each head from candidate query
features alone, with the encoder's parameters excluded from training --
drops booking PR-AUC from $0.2108$ to $0.1410$ ($-33.1\%$ relative) and
degrades every other head by a comparable or larger margin (Eng.\ Macro
$-39.5\%$, Eng.\ Micro $-34\%$, Bedrooms $-34.9\%$, Beds $-59.4\%$,
Bathrooms $-67.9\%$; Table~\ref{tab:offline}). This
regression across every task, confirms that the sequence encoder,
not the MLP heads alone, is the primary source of SIFT's improvement
over the aggregated-feature baseline.

\subsection{Cross-Domain Generalization: Hotel-Intent Case Study}\label{sec:hotel-intent}
\begin{figure}[t]
  \centering
  \includegraphics[width=0.95\columnwidth,height=0.33\textheight,keepaspectratio]{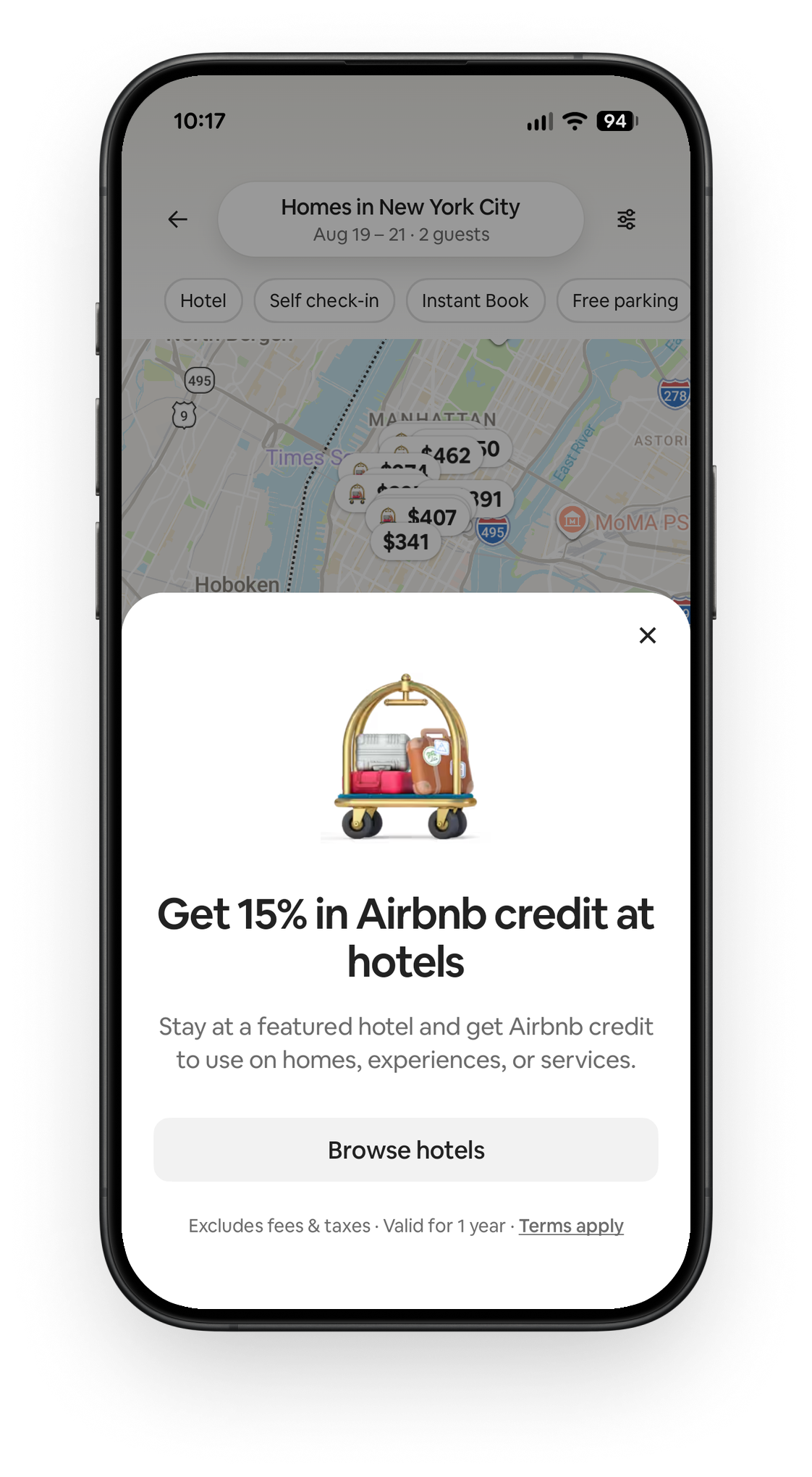}
  \caption{Hotel Megaphone on Airbnb search page}
  \Description{Screenshot of the Airbnb search page with a promotional Hotel Megaphone banner offering hotel credits.}
  \label{fig:hotel-megaphone}
\end{figure}

To improve targeting for Airbnb's \emph{Hotel Megaphone}, a promotional 
banner on search page that highlights hotel credit 
offers and routes guests into a filtered hotel search experience 
(Figure~\ref{fig:hotel-megaphone}), we extended SIFT's multi-label 
engagement head ($T2$) by adding a dedicated \textsc{Hotel} filter 
class. 
Targeting for this surface previously relied on a static, rule-based heuristic. Because the heuristic used coarse eligibility rules rather than a learned intent signal, it offered limited precision, and a more targeted signal presented a clear opportunity to improve relevance for guests on both the hotel and home surfaces.

Because a guest's historical action sequence, such as past 
hotel-filtered searches and bookings, provides strong signal of hotel intent, SIFT's user representation $\mathbf{e}_u$ computes a personalized, 
query-conditioned hotel-intent score $P(E_{\text{hotel}} \mid u, q)$. 
Because personalization is decoupled from query and filter features, this 
extension required no new feature engineering or ETL pipelines.

We evaluated the learned hotel-intent score offline within the exact 
cohort targeted by the rule-based heuristic, so the evaluation aligns 
directly with the baseline that the score is intended to replace. 
Because evaluation is conducted within this pre-filtered cohort, the 
heuristic baseline naturally has a recall of 1.0. 
Table~\ref{tab:hotel-intent} reports the heuristic precision, model 
precision, model recall, and precision lift at each signal's optimal 
operating threshold across three downstream hotel-specific interactions. 
 For our primary business conversion objective, Bookings (confirmed hotel reservations), SIFT improves precision by $2.07\times$ over the heuristic ($0.0085$ vs. $0.0041$) at a decision threshold of $0.11$, while maintaining $34\%$ recall. The two softer 
hotel-focused engagement signals (PDP Views for viewing a hotel 
property detail page, and CTA Clicks for interacting with the 
hotel promotional banner) demonstrated comparable gains. 

 We validated the hotel-intent score in an online A/B test comparing SIFT gated Hotel Megaphone targeting against both a control with the banner disabled and the existing heuristic-gated targeting.

 Table~\ref{tab:hotel-ab} reports results for the SIFT-gated treatment. Hotel-surface engagement and booking metrics all move favorably, with uncancelled Hotel bookings rising measurably, accompanied by a corresponding rise in uncancelled overall marketplace bookings ($+0.76\%$). The heuristic-gated treatment does not show this marketplace-wide gain, consistent with a static rule that may surface the banner to guests with low hotel intent, reducing offer relevance and diverting guests from listings that better match their needs. This indicates the hotel-intent score converts Hotel-surface gains into bookings that are additive to, rather than drawn from, the rest of the marketplace, whereas the heuristic's coarse targeting does not.

Together, 
these results confirm that SIFT effectively isolates high-intent users to 
replace static heuristic gating while yielding a reusable intent score for additional hotel surfaces.

\begin{table}[!t]
  \centering
  \footnotesize
  \setlength{\tabcolsep}{3.5pt}
  \caption{Offline evaluation of hotel-intent signal against heuristic. $\tau$ best-precision
  threshold; lift relative to heuristic}
  \label{tab:hotel-intent}
  \begin{tabular}{@{}lc cc cc@{}}
    \toprule
     & & \multicolumn{2}{c}{Precision} & & \\
    \cmidrule(lr){3-4}
    Signal & $\tau$ & Heuristic & SIFT & Recall & Lift \\
    \midrule
    Bookings & 0.11 & 0.0041 & 0.0085 & 0.34   & $2.07\times$ \\
    PDP Views         & 0.11 & 0.141  & 0.226  & 0.27   & $1.60\times$ \\
    CTA Clicks        & 0.09 & 0.218  & 0.293  & 0.33   & $1.34\times$ \\
    \bottomrule
  \end{tabular}
\end{table}

\begin{table}[!t]
\small
\centering
\caption{Online A/B results for hotel-intent score. First five metrics: Hotel surface;
last: marketplace-wide. All statistically significant.}
\label{tab:hotel-ab}
\begin{tabular}{lc}
\toprule
\textbf{Metric} & \textbf{Relative lift} \\
\midrule
Searchers                                 & $+16.9\%$ \\
PDP Views                                 & $+5.2\%$ \\
\textbf{Bookings Not Cancelled} & $\mathbf{+3.8\%}$ \\
Nights Booked Not Cancelled     & $+5.9\%$ \\
Gross Booking Value             & $+5.5\%$ \\
\midrule
\textbf{Bookings Not Cancelled (marketplace-wide)} & $\mathbf{+0.76\%}$ \\
\bottomrule
\end{tabular}
\end{table}
\FloatBarrier

\section{Discussion and Limitations}
\label{sec:discussion}

The primary remaining gap is same-day sessions: an estimated $\sim$30--40\% of Airbnb bookings complete within a day of the first search, too fast for SIFT's daily-refreshed user embeddings to reflect. At serving time these guests are not treated as cold-start; they are scored using $\mathbf{e}_u$ from the previous day's refresh, which does not yet incorporate that day's own searches or bookings. The practical effect is staleness rather than absence: filter predictions for same-day guests rely on the previous day's embedding rather than a missing signal. This is the same daily-refresh/latency trade-off adopted by JourneyFormer~\cite{journeyformer}, whose authors similarly note that a shorter refresh cadence would improve online performance but requires additional streaming infrastructure. We plan to pursue the same direction, prioritizing same-day sessions as the segment with the most to gain from more frequent updates.

\section{Conclusion}
\label{sec:conclusion}
SIFT replaces hand-engineered aggregated features with transformer generated user-sequence embeddings in a multi-task framework. Offline, it improves booking PR-AUC by $+51.9\%$ and amenity engagement PR-AUC by $+62.8\%$. Online, it lifts Recommended Filter Clickers by $+20\%$, Searchers With Filters by $+0.72\%$, and Bathroom/Bedroom filter usage by $+10.7\%$/$+3.9\%$. Extended to a new hotel-intent filter with only a new label and head, it lifts uncancelled hotel bookings by $+3.8\%$ and marketplace bookings by $+0.76\%$ online. SIFT is now fully deployed, serving scalable personalization to millions of guests.

\balance
\bibliographystyle{ACM-Reference-Format}

\end{document}